\documentclass[runningheads]{llncs}
\usepackage[T1]{fontenc}
\usepackage{graphicx}
\usepackage{hyperref}
\usepackage[inkscapelatex=false]{svg}
\usepackage{colortbl}
\usepackage[smartEllipses]{markdown}
\usepackage{comment}
\usepackage{multirow}
\begin{document}
\title{LLMs4SchemaDiscovery: A Human-in-the-Loop Workflow for Scientific Schema Mining with Large Language Models}
\titlerunning{A Human-in-the-Loop Workflow for Scientific Schema Mining with LLMs}
%
\author{Sameer Sadruddin\inst{1}\thanks{Both authors contributed equally to this work.} \and Jennifer D'Souza\inst{1}$^{*}$ \and Eleni Poupaki\inst{2} \and Alex Watkins\inst{3} \and Hamed Babaei Giglou\inst{1} \and Anisa Rula\inst{4} \and Bora Karasulu\inst{3} \and S\"{o}ren Auer\inst{1,5} \and Adrie Mackus\inst{2} \and Erwin Kessels\inst{2}}

\authorrunning{Sadruddin \& D'Souza et al.}
%
\institute{TIB  Leibniz Information Centre for Science and Technology, Hannover, Germany \and
TU/e Eindhoven University of Technology, Netherlands \and
University of Warwick, United Kingdom
\and
University of Brescia, Italy
\and
L3S Research Center, Leibniz University of Hannover, Germany\\
\email{\{sameer.sadruddin,jennifer.dsouza\}@tib.eu}}
\maketitle              
\begin{abstract}
Extracting structured information from unstructured text is crucial for modeling real-world processes, but traditional schema mining relies on semi-structured data, limiting scalability. This paper introduces \textsc{schema-miner}, a novel tool that combines large language models with human feedback to automate and refine schema extraction. Through an iterative workflow, it organizes properties from text, incorporates expert input, and integrates domain-specific ontologies for semantic depth. Applied to materials science—specifically atomic layer deposition—\textsc{schema-miner} demonstrates that expert-guided LLMs generate semantically rich schemas suitable for diverse real-world applications.


\end{abstract}

\begin{itemize}
    \item[] \textbf{Resource type}: Software
    \item[] \textbf{License}: MIT License
    \item[] \textbf{DOI}: \url{https://doi.org/10.5281/zenodo.14781824}
    \item[] \textbf{URL}: \url{https://github.com/sciknoworg/schema-miner}
\end{itemize}

\section{Introduction}

Scientific research generates vast amounts of structured, semi-structured, and unstructured data across domains like biology, chemistry, physics, and environmental science. Although findings are mainly communicated through unstructured research papers, these documents offer limited structured metadata and often lack formal mechanisms to capture content-level information. Yet, the expertise embedded in such papers makes them a rich source of scientific knowledge, with strong potential for structured pattern discovery. Since the goal and challenges of schema information extraction vary with the input type, content, and desired output, this paper contributes a software resource for extracting schemas from natural language sources as a foundation for broader knowledge applications.

Schemas are essential for standardizing data, enabling validation, ensuring interoperability, and supporting tasks like knowledge extraction, data integration, and automation \cite{bizer2023linked,sheth1990federated,beeri1999schemas,adam_shepherd_2023}. They form the backbone of ontology engineering (OE) \cite{guarino2009ontology,noy2001ontology} and knowledge graph (KG) construction \cite{ye2023schema,zhou2022schere}, especially in scientific domains. For example, the Open Research Knowledge Graph (ORKG) \cite{auer2020improving} leverages expert-defined, domain-specific schemas to enhance the discoverability and integration of scientific knowledge. However, schema development in science remains fragmented, domain-specific, and often manually crafted, lacking generalizability. Much scientific data lacks formal schemas or follows undocumented formats, making schema discovery—a critical step for standardization and automation—especially challenging. Addressing this requires tools that can formalize complex data structures. Large Language Models (LLMs), with their strengths in pattern recognition and text synthesis, offer a promising solution.

LLMs \cite{llm-amatriain-arxiv,llm-dsouza-orkg} are well-suited for schema discovery due to their ability to recognize patterns, synthesize knowledge, and generate human-like outputs. 
General-purpose LLMs \cite{achiam2023gpt,openai2024gpt4,touvron2023llama,llama2024,jiang2023mistral,mistral2024,bai2023qwen,qwen2024} are known to be pretrained on large code corpora, enabling them to learn meaningful representations that support extracting text- and code-based schemas. 
Their ability to capture structural and semantic patterns makes them effective for generating standardized schema formats like JSON from unstructured data. However, human expertise remains essential to ensure accuracy and domain relevance, highlighting the value of human-in-the-loop workflows that combine LLM output with expert refinement.

In this resource paper, we introduce \textbf{\textsc{schema-miner}}, a first-of-its-kind software tool for schema mining in scientific domains. Our approach, called LLMs4SchemaDiscovery, leverages LLMs to identify candidate schemas from small to large scientific paper collections. Designed in a \textbf{plug-and-play} modus, it supports seamless integration with most LLMs via the \href{https://ollama.com/}{Ollama} or \href{https://huggingface.co/models}{HuggingFace} libraries and accepts user-defined literature collections—making it adaptable across models and domains. The approach follows a three-stage workflow: \textit{initial schema design}, \textit{refinement}, and \textit{finalization}, with the latter two incorporating \textbf{human-in-the-loop} feedback to ensure schema quality and domain relevance. By combining LLM pattern recognition with expert oversight, \textsc{schema-miner} accelerates schema development and promotes scientific data standardization. 
The paper makes three key contributions. \textbf{1. Systematic Methodology:} A scalable, structured approach for schema discovery using LLMs. \textbf{2. Human-in-the-Loop Workflow:} An adaptable pipeline that combines automated extraction with expert refinement. 
\textbf{3. Practical Demonstration:} We apply \textsc{schema-miner} to a materials science use case on Atomic Layer Deposition (ALD), demonstrating its ability to extract meaningful schemas for generating AI-ready KGs that enable machine-actionable, integrative research across industrially relevant scenarios.

\section{Related Work}

Schema discovery involves identifying the underlying structure of data and has evolved significantly due to the increasing demand for data integration and understanding. Early approaches focused on syntactic characteristics and statistical analysis of tabular data, leveraging predefined rules, frequency distributions, and clustering techniques to identify schema elements. However, these methods often struggled with semantic understanding, handling noisy data, and required manual intervention for maintaining rules or labeled training data \cite{MillerA03,LiuCTHLM23}. Machine learning approaches, particularly deep learning, have been applied to schema matching tasks, using clustering techniques to group similar data values (e.g., ``male'' and ``female'') and neural networks to learn intricate patterns and relationships. While these methods enhance schema matching accuracy \cite{hu2019survey,paulheim2020automating,aldafian2021deep}, they differ from our task of schema discovery as they focus on aligning existing schemas rather than identifying new structures, and they often require large labeled datasets with limited interpretability. Another related line of work is semantic schema enrichment, which builds on initial schema discovery, which we do in this work, by linking data to existing knowledge. It includes inferring schema elements from implicit data structures, such as graph-based datasets \cite{Kirchberg2012,Lutov2018}, enriching schemas with new statements or entity typing (e.g., \texttt{rdf:type}) \cite{BiswasPPSA22,Paulheim2013}, and identifying schema patterns like frequent RDF graph structures \cite{Kirchberg2012,BiswasPPSA22,BaaziziCGS19}. The rise of NoSQL databases has introduced new challenges for schema discovery due to their lack of predefined schemas and high variability in data representation. Work has focused on inferring implicit schemas from NoSQL data, such as using hierarchical clustering for property graphs \cite{BonifatiDMGJLP22} and schema inference algorithms for JSON documents \cite{baazizi2017schema,canovas2013discovering}. More recent approaches leverage LLMs to extract attribute values and add semantic meaning to discovered schemas, complementing traditional techniques like clustering and rule-based methods \cite{brinkmann2023extractgpt,mior2024large}.

Unlike prior work focused on schema alignment, enrichment, or domain-specific inference, our approach introduces a generalizable, human-in-the-loop workflow for schema discovery using LLMs. It uniquely combines automated extraction with expert refinement to address domain specificity, semantic complexity, and data fragmentation. Implemented in the \textsc{schema-miner} tool, this is the first framework to systematically mine schemas from unstructured scientific text at scale, adaptable to various LLMs and domains. This sets our work apart from recent LLM-based efforts that emphasize schema enrichment \cite{mior2024large} rather than full schema discovery.


\section{Task Definition}

This work addresses the task of \textit{schema discovery}—identifying and formalizing the underlying structure of data in scientific papers. Schema discovery is key to standardizing diverse scientific data and supporting automation, integration, and formal reasoning. A \textit{schema} \( S \) is defined as a tuple: $S = (P, T, C, R)$, where \( P \) is a set of \textit{properties}, \( T: P \rightarrow \mathcal{T} \) maps each property to a data type \( \mathcal{T} \), \( C \) defines \textit{constraints}, and \( R \) captures \textit{relationships} among nested structures. The goal is to extract candidate schemas \( S_* \) from unstructured data \( D = \{d_1, d_2, \dots, d_n\} \), where \( d_i \in \mathcal{D} \) and \( \mathcal{D} \) includes text-based or semi-structured sources.

\section{Our Approach: LLMs4SchemaDiscovery}
\label{sec:approach}

\begin{figure}[!htb]
    \centering
    \includegraphics[width=\linewidth]{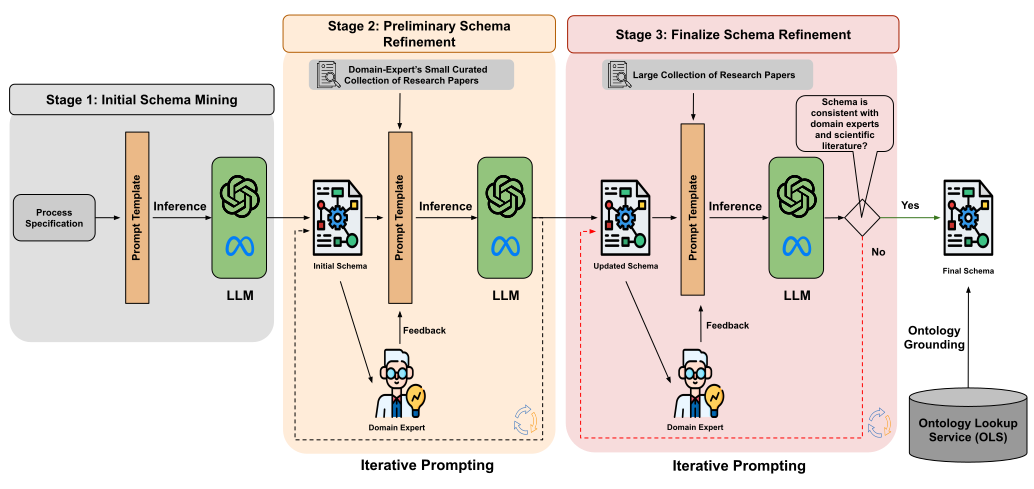}
    \caption{Overview of the LLMs4SchemaDiscovery workflow implemented in \href{https://github.com/sciknoworg/schema-miner}{\textsc{schema-miner}}. Stage 1 (gray box) generates an initial schema from domain specifications. Stage 2 (orange box) refines the schema using a small, expert-curated set of papers and optional feedback. Stage 3 (red box) finalizes the schema with a larger, non-curated collection of papers. The workflow iteratively updates the schema and concludes by grounding schema properties to ontologies using an ontology lookup service API.}
    \label{fig:workflow}
\end{figure}

The LLMs4SchemaDiscovery framework (see \autoref{fig:workflow}) comprises three iterative stages: initial schema generation, refinement, and finalization. The process begins with a specification document provided as input to the LLM, which generates an initial schema. This schema is iteratively refined using curated scientific knowledge from papers and expert feedback, enhancing its structural and semantic relationships. The final schema offers a comprehensive, semantically enriched representation of the target domain, with mechanisms to organize schema properties and relationships through an ontology lookup service API.

\subsection{Stage 1: Generate Initial Schema}

In this stage, the LLM generates an initial JSON schema encompassing essential properties, their data types, and associated constraints for the target domain. This foundational schema, guided by structured prompts and domain specifications, serves as the basis for further refinement and enrichment with domain-specific knowledge and ontological relationships in subsequent stages. The process is formalized as follows:

\textbf{GenerateSchema}$(P, T, C, R)$ is defined as 
\begin{multline*}
    \href{https://github.com/sciknoworg/schema-miner/blob/main/src/prompts/prompt_template1.py}{\textit{SystemPrompt}}(\text{Role}, \text{Task}, \text{InputFormat}, \text{OutputFormat}) + \\
    \href{https://github.com/sciknoworg/schema-miner/blob/main/src/prompts/prompt_template1.py}{\textit{UserPrompt}}(\text{DomainSpecifications}),
\end{multline*}
where \textit{SystemPrompt} specifies structured instructions provided to the LLM, consisting of: 
\textbf{Role}, assigning the LLM a specific function (e.g., ``You are an expert in schema design for scientific data''); 
\textbf{Task}, defining the goal (e.g., ``Identify properties, data types, constraints, and relationships to generate an initial schema from the provided domain specifications''); 
\textbf{InputFormat}, describing the format of the input data that will be provided via the user prompt, such as free text or semi-structured data; and 
\textbf{OutputFormat}, defining the expected schema structure—including $P$, $T$, $C$, and $R$—formatted as JSON.
\textit{UserPrompt}$(\text{DomainSpecification})$ represents the unstructured or semi-structured domain specifications provided by the human expert, including descriptive details, domain-specific examples, or predefined metadata. 
The output is the initial schema \( S_1 = (P_1, T_1, C_1, R_1) \), where \( P_1 \) denotes properties identified by the LLM, \( T_1 \) maps data types to properties, \( C_1 \) includes inferred or specified constraints, and \( R_1 \) captures relationships among nested structures.

\subsection{Stage 2: Refine Schema} 

The refinement stage enhances the initial schema by iteratively analyzing a curated collection of scientific papers and incorporating expert feedback. Using a small set of 1 to 10 expert-selected papers, the LLM refines the schema by updating its properties, mapping refined data types, revising constraints, and identifying or modifying relationships between schema elements. The iterative process ensures the schema becomes both specific and generalizable, capturing structural and semantic consistency across various research descriptions. A human-in-the-loop approach is employed, where domain experts review the schema at each iteration, providing feedback to guide updates and address potential errors or omissions. 

\textbf{RefineSchema}$(P, T, C, R)$ is defined as 
\begin{multline*}
    \href{https://github.com/sciknoworg/schema-miner/blob/main/src/prompts/prompt_template2.py}{\textit{SystemPrompt}}(\text{Role}, \text{Task}, \text{InputFormat}, \text{OutputFormat}) + \\
    \href{https://github.com/sciknoworg/schema-miner/blob/main/src/prompts/prompt_template2.py}{\textit{UserPrompt}}(\text{PrevSchema}, \text{SciPaper}, \text{ExpertFeedback}),
\end{multline*}
where \textit{SystemPrompt} specifies the structured instructions provided to the LLM and consists of: 
\textbf{Role}, which defines the LLM's function (e.g., ``You are an expert in refining scientific schemas using curated papers and feedback''); 
\textbf{Task}, which specifies the objective (e.g., ``Iteratively refine the schema by analyzing properties, data types, constraints, and relationships within the context of the provided scientific paper and feedback, while ensuring semantic consistency''); 
\textbf{InputFormat} and \textbf{OutputFormat} are the same as before. 

The \textit{UserPrompt} incorporates \textbf{PrevSchema}, i.e., the schema $S_1$ generated in Stage 1 or the previous step in Stage 2. \textbf{SciPaper}: A single paper from a curated collection of $n$ scientific papers, where $1 \leq n \leq 10$. Each paper provides domain-specific information to enrich the schema. \textbf{ExpertFeedback} (optional): Feedback from a domain expert to further improve the schema. 

Expert feedback in this stage is solicited based on clear guidelines to ensure relevance and clarity. Experts can provide feedback in one of two ways: (1) as descriptive text addressing four guiding questions—Should any properties be merged, and what would you name the merged property? Which properties should be grouped into a single unit, and how would you describe it? Are there any essential properties missing? Are the current property descriptions clear and comprehensive? Or (2) through direct edits to the schema, modifying properties, constraints, or relationships as needed. These methods balance high-level conceptual feedback with precise, actionable changes. Our \href{https://github.com/sciknoworg/schema-miner/blob/main/assets/LLMs4SchemaDiscovery%20-%20Domain%20Expert%20Feedback%20Guidelines.pdf}{domain expert feedback guidelines} is released online. By incorporating both feedback types, this stage iteratively refines the schema while evaluating the most effective method for LLM-driven improvements. Systematic empirical tests are explored in \autoref{sec:results}.

This iterative process continues for each paper in the curated set, progressively refining the schema through semantic enrichment, expert guidance, and LLM-driven observations from the research papers, ensuring the schema aligns with the domain's structural and semantic requirements.

The output of this stage is the refined schema \( S_2 = (P_2, T_2, C_2, R_2) \), where \( P_2 \) represents the updated set of properties derived from the scientific paper and expert feedback, \( T_2 \) maps these refined properties to their updated data types, \( C_2 \) captures the revised constraints inferred or specified during the refinement process, and \( R_2 \) encompasses any new relationships between schema elements.

\subsection{Stage 3: Finalize Schema}

The finalization stage refines the schema using a larger, uncurated corpus of up to 100 or more scientific papers to enhance its structural and semantic comprehensiveness and generalizability. The LLM validates and expands the schema by incorporating new properties, correcting omissions, and ensuring appropriate data types and constraints while avoiding irrelevant or redundant additions. Domain-expert feedback remains critical, guiding iterative refinements to ensure accuracy, generalizability, and semantic robustness.

\textbf{FinalizeSchema}$(P, T, C, R)$ is defined as 
\begin{multline*}
   \href{https://github.com/sciknoworg/schema-miner/blob/main/src/prompts/prompt_template3.py}{\textit{SystemPrompt}}(\text{Role}, \text{Task}, \text{InputFormat}, \text{OutputFormat}) + \\
    \href{https://github.com/sciknoworg/schema-miner/blob/main/src/prompts/prompt_template3.py}{\textit{UserPrompt}}(\text{PrevSchema}, \text{SciPaper}, \text{ExpertFeedback}),
\end{multline*}
where the structure follows \textbf{RefineSchema} but adapts to the larger dataset. 

The output of this stage is the finalized schema \( S_3 = (P_3, T_3, C_3, R_3) \), representing a comprehensive and semantically robust structure for the target process. This stage ensures the schema's completeness and general applicability across diverse research descriptions.

Contrasting stages 2 and 3, the refinement stage 2, which uses a small, domain-expert curated papers (1–10), helps establish a high-quality, domain-grounded baseline schema. By first aligning the LLM-generated schema with authoritative, carefully selected example research papers, the model can integrate core concepts and domain-specific nuances more reliably. This ensures that the schema’s foundational structure and terminology are accurate before it is exposed to the larger, more heterogeneous collection of up to 100 or even more non-curated papers in stage 3. As a result, the subsequent broader refinement stage can generalize and expand the schema while retaining its integrity and relevance, ultimately leading to a more robust and widely applicable schema.

\subsection{Stage 4: Ontology Grounding} 

The ontology lookup service (OLS) represents the final stage of schema discovery, grounding schema elements to relevant ontology concepts. Designed with a plug-and-play modus, the OLS can integrate with APIs from various institutions for their curated knowledge bases tailored to specific domains. This flexibility allows the system to adapt to different scientific disciplines. OLS integration into \textsc{schema-miner} involves four steps: 1) Preprocessing: Replace underscores with spaces for ontology label compatibility. 2) Search: Query OLS resources—classes, properties, individuals, and ontologies—via the OLS API. 3) Validation: Exclude candidates lacking descriptions to ensure interpretability. 4) Ranking: Use a \href{https://huggingface.co/sentence-transformers}{sentence-transformer model} to score similarity between query terms and ontology descriptions, prioritizing description-based matches to select top-k candidates. This process enhances schema properties with ontology-grounded elements, ensuring semantic rigor and alignment with domain standards.

For instance, in the materials science domain, the \href{https://terminology.tib.eu/ts}{TIB Terminology Service} (TS) offers seamless integration as a curated knowledge base of ontologies, supporting semantic alignment. Of the 94 ontologies identified in \cite{norouzi2024landscape}, 21 are available in TS, along with 19 from \cite{stromert2022ontologies4chem}. The \href{https://terminology.tib.eu/ts/collections}{NFDI4Chem project} collection significantly overlaps with these, comprising 39 ontologies. To address gaps, four additional ontologies—`ChemOnt,' `OntoKin,' `QUDT,' and `OntoCAPE'—were manually added, resulting in a refined pool of 50 unique ontologies. By integrating metadata from TS, the lookup service enriches schema properties with ontological context, enhancing semantic coherence and interoperability.

\section{\textsc{schema-miner} Tool Usage}

The \textsc{schema-miner} tool enables users to discover schemas in JSON format for a target domain in a practical and configurable manner. Parameters such as API keys, base URLs, and model settings are managed in a dedicated \texttt{.env} \href{https://github.com/sciknoworg/schema-miner/blob/main/.env.example}{configuration file}. The workflow begins by preparing a knowledge base, which includes an initial domain specification document and relevant research papers. For machine processing, the input documents are in plain text format. If research papers are in PDF format, \textsc{schema-miner} includes a script that converts them to plain text using LangChain's \href{https://python.langchain.com/api_reference/community/document_loaders/langchain_community.document_loaders.pdf.PyPDFLoader.html}{\texttt{PyPDFLoader}}. The script takes the directory of the PDFs and returns the text files to the same directory.

Schema mining proceeds in three stages after preparing the knowledge base. \textbf{Stage 1} generates an initial JSON schema using the domain specification and selected LLM.  
\texttt{Input}: Domain specification and LLM configuration.  
\texttt{Output}: Initial JSON schema saved to the specified directory.
\textbf{Stage 2} iteratively refines the schema by analyzing curated research papers and integrating expert feedback.  
\texttt{Input}: Initial schema, curated papers, and expert input.  
\texttt{Output}: Iteratively updated JSON schema.
\textbf{Stage 3} validates and finalizes the schema using a broader, uncurated corpus to ensure generalizability and semantic robustness.  
\texttt{Input}: Refined schema and uncurated research papers.  
\texttt{Output}: Final JSON schema with improved coverage.
A human-in-the-loop workflow supports user feedback and iteration control. Sample runs and detailed documentation are in the \href{https://github.com/sciknoworg/schema-miner/blob/main/README.md}{README}.

\section{Application: Materials Science Use Case}

In 2023, industry leaders \href{https://www.merckgroup.com/en}{Merck} and \href{https://www.intel.de/content/www/de/de/homepage.html}{Intel} launched the ``AI-Aware Pathways to Sustainable Semiconductor Process and Manufacturing Technologies (AWASES)'' project \cite{merck_intel_ai_project}, in the Materials Sciences with a key focus on, ``Turning Online Atomic Layer Deposition (ALD) and Etching (ALE) Databases Into AI-Ready Tools for Development of New Sustainable Materials and Fabrication Processes'' \cite{mackus2024ald}. ALD’s precise control over thin-film deposition is critical for advancing high-performance semiconductor and electronics manufacturing, making it a priority for these industry leaders. To address this, we propose semantic modeling of ALD processes in KGs as a solution for creating AI-ready representations. Schema discovery is a crucial step in this endeavor, and we take ALD processes as a use case to test \textsc{schema-miner}. Specifically, we apply \textsc{schema-miner} to generate schemas for ALD process descriptions in two contexts: experimental and simulation. This use case is particularly compelling due to the involvement of domain experts within the project collaboration, who provide iterative feedback on the generated schemas during refinement stages 2 and 3. The remainder of the paper presents systematic experiments in both ALD settings, demonstrating the efficacy of \textsc{schema-miner} as a schema discovery tool.

\begin{figure}[!htb]
    \centering
    \includegraphics[width=0.6\linewidth]{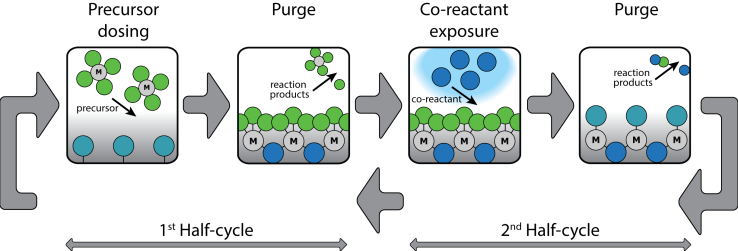}
    \caption{An ALD cycle of two half-reactions: precursor addition and surface reaction, followed by co-reactant intro., with purge phases ensuring clean, controlled growth \cite{vos2019atomic}.}
    \label{fig:ALD-cycle}
\end{figure}

The reproducibility and control of ALD processes (see \autoref{fig:ALD-cycle}) make them essential for advanced electronics, photovoltaics, energy storage, and catalysis \cite{alvaro2018characterizing}, while their consistency supports predictive AI models for process development. ALD is performed through laboratory experiments and computational simulations. \textbf{ALD experiments}, though based on straightforward self-limiting chemical processes, require meticulous design and execution to ensure reliable outcomes \cite{shahmohammadi2022recent,vos2019atomic}. Researchers must define the application, select suitable precursors, co-reactants, and substrates, and optimize parameters such as pulse and purge durations, reactor type, temperature, and pressure. Repeated experiments ensure reproducibility, and characterization verifies properties like thickness, growth per cycle, structure (e.g., crystallinity, composition), and morphology (e.g., conformality, uniformity) \cite{knoops2015atomic}. In contrast, \textbf{ALD simulations} advance process understanding, complementing experiments or serving as standalone studies. These range from atomistic models, which explore mechanisms and reaction energies \cite{Mameli2017}, to continuum models analyzing reactor-scale properties like gas flow \cite{Yanguas-Gil2021}. Commonly studied properties, such as reaction step energies or growth rates \cite{Sengupta2005}, often lack standardized reporting in scientific papers, hindering analysis and comparison. Semantic web technologies can address this by structuring key information, improving the efficiency of literature searches and studies for ALD.

\section{Experiments and Results}
\label{sec:results}

We now present our experiments applying \textsc{schema-miner} to ALD process experiments and simulations to discover their schemas, respectively. In stage 1, domain experts provided \href{https://github.com/sciknoworg/schema-miner/tree/main/data/stage-1}{process specification documents} for both ALD experiment and simulation scenarios. In stage 2, the experts curated \href{https://github.com/sciknoworg/schema-miner/tree/main/data/stage-2/research-papers}{seven high-quality papers} for each setting. Finally, in stage 3, the \href{https://github.com/sciknoworg/schema-miner/tree/main/data/stage-3/research-papers}{research papers collection} was sourced from the \href{https://www.atomiclimits.com/alddatabase/}{AtomicLimits} ALD Database developed by TU/e in 2019. Next, we provide details of the experimental setup.

\subsection{Experimental Setup}

The \textsc{schema-miner} tool is implemented in Python, using the \href{https://python.langchain.com/docs/introduction/}{LangChain} library for interfacing with LLMs. In our experiments, we tested three models: GPT-4o (ver. 2024-08-06), GPT-4-turbo (ver. 2024-04-09), and LLama 3.1 (8B). For OpenAI models (GPT-4o and GPT-4-turbo), we used the \href{https://python.langchain.com/api_reference/openai/chat_models/langchain_openai.chat_models.base.ChatOpenAI.html}{ChatOpenAI} class from LangChain to interface with OpenAI services. For LLama 3.1 (8B), we utilized the Scalable AI Accelerator (SAIA) platform, which supports multiple open-source LLMs via an OpenAI-compatible API, ensuring seamless integration with tools like ChatOpenAI. \textsc{schema-miner} also integrates with \href{https://github.com/ollama/ollama}{Ollama}, providing users a choice of a wide range of open-source LLMs. The key hyperparameters adjusted in our experiments were context length and temperature. Context length varied based on model architecture; for the three models used in our study, we set a uniform context length of 128K tokens. To balance output stability and creativity, we used a fixed temperature value of 0.3 across all models. For the cloud-based GPT models, \textsc{schema-miner} was executed using minimal local resources on a personal laptop. Although the Llama 3.1 8B model is CPU-compatible, we leveraged an institutional GPU node (64-core CPU, 500 GB RAM, RTX 3090 GPU) to expedite inference. Notably, the model itself requires no more than 16 GB of RAM to run. A key feature of \textsc{schema-miner} is its flexibility and suitability, especially in compute-constrained environments, allowing users to integrate their LLM of choice, including smaller distilled or quantized LLM variants. The compute requirements scale with the model size and the applied quantization level.

Earlier in \autoref{sec:approach}, we introduced two methods for soliciting domain expert feedback: (1) descriptive text and (2) direct edits to the schema. These methods informed the design of our experimental settings, which comprised four main types. In \textbf{Experiment 1}, descriptive feedback was provided in two variants: (a) included only at the first iteration or (b) included at every iteration. In \textbf{Experiment 2}, expert-edited schemas were used, with reviews conducted either (a) only at the first iteration or (b) at every iteration. \textbf{Experiment 3} combined both descriptive feedback and expert-edited schemas, also with two variants: (a) reviewed only at the first iteration or (b) reviewed at every iteration. Finally, \textbf{Experiment 4} served as a baseline with no feedback. These variations aimed to evaluate the model's sensitivity to different feedback types and their impact on schema quality. Experiment 3, where both feedback formats were incorporated at every iteration, emerged as the best-performing protocol based on domain expert evaluations, identifying its schema as the most accurate representation of the ALD process. A total of 21 experiments were conducted with \textsc{schema-miner} end-to-end, comprising seven experiments for the three LLMs tested. 

\subsection{Results and Discussion}

As a first step, we conduct a surface analysis of the variability in the schemas generated by the three LLMs across stages 1, 2, and 3 w.r.t. quantitative measures for just the schemas from Experiment 3.

\begin{table}[!htb]
\centering
\caption{Quantitative schema variance across Stages 1, 2, and 3 of \textsc{schema-miner} for ALD experimental processes, evaluated using ROUGE-L, BLEU, and BERTScore metrics, comparing schemas from GPT-4o, GPT-4-turbo, and LLama 3.1 (8B).}
\label{tab:quantitative-results}
\resizebox{\textwidth}{!}{%
\begin{tabular}{cccccccccc}
\hline
\multicolumn{10}{c}{\textbf{Stage 1}} \\ \hline
\multicolumn{1}{c|}{} & \multicolumn{3}{c|}{\textbf{GPT-4o}} & \multicolumn{3}{c|}{\textbf{GPT-4-turbo}} & \multicolumn{3}{c}{\textbf{LLama-3.1-8B}} \\ \cline{2-10} 
\multicolumn{1}{c|}{} & \textbf{RougeL} & \textbf{Bleu Score} & \multicolumn{1}{c|}{\textbf{BERT-F1}} & \textbf{RougeL} & \textbf{Bleu Score} & \multicolumn{1}{c|}{\textbf{BERT-F1}} & \textbf{RougeL} & \textbf{Bleu Score} & \textbf{BERT-F1} \\
\multicolumn{1}{c|}{\textbf{GPT-4o}} & \cellcolor[HTML]{C0C0C0} & \cellcolor[HTML]{C0C0C0} & \multicolumn{1}{c|}{\cellcolor[HTML]{C0C0C0}} & 0.3428 & 0.4022 & \multicolumn{1}{c|}{0.8098} & 0.3100 & 0.2862 & 0.8044 \\
\multicolumn{1}{c|}{\textbf{GPT-4-turbo}} & 0.3428 & 0.3916 & \multicolumn{1}{c|}{0.8098} & \cellcolor[HTML]{C0C0C0} & \cellcolor[HTML]{C0C0C0} & \multicolumn{1}{c|}{\cellcolor[HTML]{C0C0C0}} & 0.4118 & 0.3481 & 0.7765 \\
\multicolumn{1}{c|}{\textbf{LLama-3.1-8B}} & 0.3100 & 0.2649 & \multicolumn{1}{c|}{0.8044} & 0.4118 & 0.3443 & \multicolumn{1}{c|}{0.7765} & \cellcolor[HTML]{C0C0C0} & \cellcolor[HTML]{C0C0C0} & \cellcolor[HTML]{C0C0C0} \\ \hline
\multicolumn{10}{c}{\textbf{Stage 2}} \\ \hline
\multicolumn{1}{c|}{} & \multicolumn{3}{c|}{\textbf{GPT-4o}} & \multicolumn{3}{c|}{\textbf{GPT-4-turbo}} & \multicolumn{3}{c}{\textbf{LLama-3.1-8B}} \\ \cline{2-10} 
\multicolumn{1}{c|}{} & \textbf{RougeL} & \textbf{Bleu Score} & \multicolumn{1}{c|}{\textbf{BERT-F1}} & \textbf{RougeL} & \textbf{Bleu Score} & \multicolumn{1}{c|}{\textbf{BERT-F1}} & \textbf{RougeL} & \textbf{Bleu Score} & \textbf{BERT-F1} \\
\multicolumn{1}{c|}{\textbf{GPT-4o}} & \cellcolor[HTML]{C0C0C0} & \cellcolor[HTML]{C0C0C0} & \multicolumn{1}{c|}{\cellcolor[HTML]{C0C0C0}} & 0.3071 & 0.4515 & \multicolumn{1}{c|}{0.8094} & 0.3535 & 0.3316 & 0.8112 \\
\multicolumn{1}{c|}{\textbf{GPT-4-turbo}} & 0.3071 & 0.4501 & \multicolumn{1}{c|}{0.8094} & \cellcolor[HTML]{C0C0C0} & \cellcolor[HTML]{C0C0C0} & \multicolumn{1}{c|}{\cellcolor[HTML]{C0C0C0}} & 0.3363 & 0.2803 & 0.7695 \\
\multicolumn{1}{c|}{\textbf{LLama-3.1-8B}} & 0.3535 & 0.3319 & \multicolumn{1}{c|}{0.8112} & 0.3363 & 0.2825 & \multicolumn{1}{c|}{0.7695} & \cellcolor[HTML]{C0C0C0} & \cellcolor[HTML]{C0C0C0} & \cellcolor[HTML]{C0C0C0} \\ \hline
\multicolumn{10}{c}{\textbf{Stage 3}} \\ \hline
\multicolumn{1}{c|}{} & \multicolumn{3}{c|}{\textbf{GPT-4o}} & \multicolumn{3}{c}{\textbf{GPT-4-turbo}} & \multicolumn{3}{c}{\textbf{LLama-3.1-8B}} \\ \cline{2-10} 
\multicolumn{1}{c|}{} & \textbf{RougeL} & \textbf{Bleu Score} & \multicolumn{1}{c|}{\textbf{BERT-F1}} & \textbf{RougeL} & \textbf{Bleu Score} & \multicolumn{1}{c|}{\textbf{BERT-F1}} & \textbf{RougeL} & \textbf{Bleu Score} & \textbf{BERT-F1} \\
\multicolumn{1}{c|}{\textbf{GPT-4o}} & \cellcolor[HTML]{C0C0C0} & \cellcolor[HTML]{C0C0C0} & \multicolumn{1}{c|}{\cellcolor[HTML]{C0C0C0}} & 0.3690 & 0.4151 & \multicolumn{1}{c|}{0.8046} & 0.3337 & 0.3397 & 0.7716 \\
\multicolumn{1}{c|}{\textbf{GPT-4-turbo}} & 0.3690 & 0.4151 & \multicolumn{1}{c|}{0.8046} & \cellcolor[HTML]{C0C0C0} & \cellcolor[HTML]{C0C0C0} & \multicolumn{1}{c|}{\cellcolor[HTML]{C0C0C0}} & 0.2891 & 0.2392 & 0.7560 \\
\multicolumn{1}{c|}{\textbf{LLama-3.1-8B}} & 0.3337 & 0.3493 & \multicolumn{1}{c|}{0.7716} & 0.2891 & 0.2458 & \multicolumn{1}{c|}{0.7560} & \cellcolor[HTML]{C0C0C0} & \cellcolor[HTML]{C0C0C0} & \cellcolor[HTML]{C0C0C0} \\ \hline
\end{tabular}%
}
\end{table}

\subsubsection{Quantitative Results.}

Here, the focus was on measuring property variance and changes across models and stages, addressing questions like: How did the schema evolve across stages? And how closely aligned were the schemas generated by three different LLMs? Due to space constraints, this discussion is limited to ALD experimental schemas. However, we provide all generated schemas—per stage, per step in stages 2 and 3, per experimental setting, and per LLM for both ALD experiments and simulations—in our \href{https://github.com/sciknoworg/schema-miner/tree/main/results}{software repository}. We measured schema variance using three complementary metrics commonly used in text generation to compare candidate and reference texts. Here, one LLM's output serves as the candidate schema compared against the outputs of the other two LLMs as reference schemas. The metrics used are ROUGE \cite{lin2004rouge}, BLEU \cite{papineni2002bleu}, and BERTScore \cite{zhang2019bertscore}. ROUGE measures recall and n-gram overlap, BLEU evaluates precision in text summarization, and BERTScore, using BERT-type embeddings, assesses semantic similarity. For BERTScore, we used the SciBERT model \cite{beltagy2019scibert}, given its suitability for scientific text. The results are shown in \autoref{tab:quantitative-results}.

In stage 1, RougeL scores highlight differences in schema alignment across LLMs. GPT-4-turbo achieved a RougeL of \textbf{0.4118} compared to LLama 3.1 (8B), indicating high alignment, while LLama 3.1 (8B) scored \textbf{0.3100} against GPT-4o, showing weaker agreement. GPT-4o had balanced performance with a RougeL of \textbf{0.3428} compared to GPT-4-turbo, suggesting greater structural similarity between the GPT-4 models than with LLama 3.1 (8B). Semantic similarity, measured by BERTScore, was consistent across models, ranging from \textbf{0.8044 to 0.8098}. In stage 2, GPT-4-turbo demonstrated strong semantic alignment with GPT-4o, achieving a BLEU score of \textbf{0.4515}, while LLama 3.1 (8B) scored \textbf{0.3316} in the same comparison. These results suggest that GPT-4-turbo produced semantically rich but slightly less structurally coherent schemas compared to LLama 3.1 (8B). By stage 3, GPT-4-turbo and GPT-4o showed strong semantic and structural alignment, with a BLEU score of \textbf{0.4151} and BERTScore of \textbf{0.8046}. Overall, the results highlight GPT-4o's robustness and adaptability, LLama 3.1 (8B)'s strength in semantic comprehension, and GPT-4-turbo's challenges in maintaining generalizability in later stages.

\subsubsection{Qualitative Results and Discussion.} This section forms the core of the paper, discussing key observations on applying \textsc{schema-miner} w.r.t. qualitative aspects of schema generation, often informed by domain expert feedback. 

\begin{figure}[!htb]
    \centering
    \includegraphics[width=\linewidth]{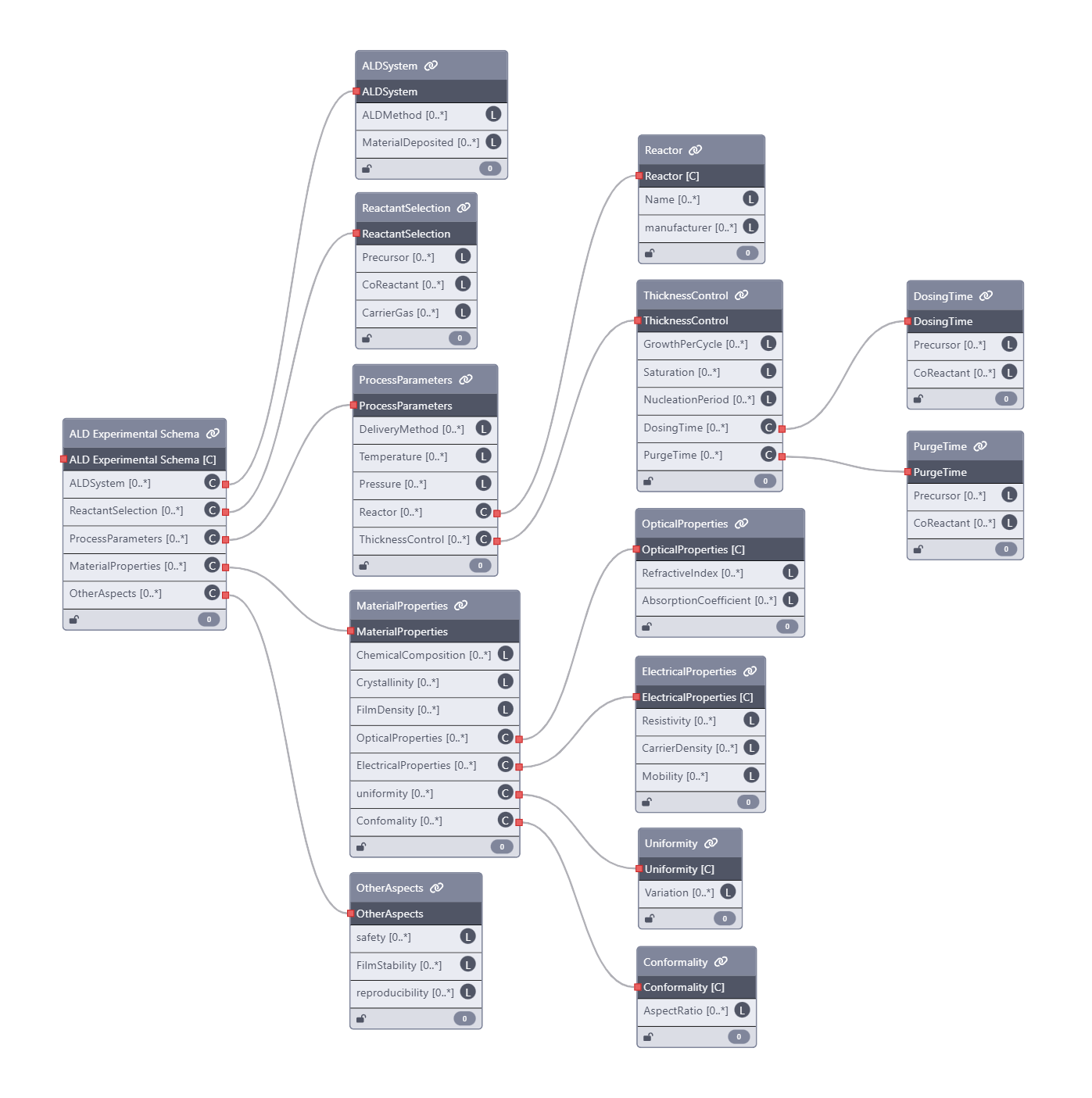}
    \caption{A UMLS diagram of the best \href{https://orkg.org/template/R796110}{ALD experimental schema} from \textsc{schema-miner}.}
    \label{fig:experimental-schema-uml}
\end{figure}

\textbf{1. LLM Stability:} Stability refers to an LLM's ability to maintain consistency across runs and avoid unnecessary additions during refinement, particularly in stages 2 and 3. GPT-4o and LLama 3.1 (8B) demonstrated high stability, introducing no irrelevant changes. In contrast, GPT-4-turbo frequently added overly specific details, reducing the schema's generalizability and utility for domain experts. \textbf{2. Comprehension of the ALD Process:} A key factor was evaluating how well each model understood the ALD process and adapted the schema accordingly. GPT-4o and LLama 3.1 (8B) demonstrated superior comprehension, capturing relationships and constraints inherent to ALD processes. While GPT-4-turbo produced acceptable schemas, its understanding was less robust. This suggests that certain LLMs may be better suited for comprehending scientific domains, which is crucial for generating high-quality schemas. \textbf{3. Complexity of the Schema:} During schema refinement in stages 2 and 3, experts observed that GPT-based models often created overly nested structures with unnecessary complexity and repetitive properties. Some schemas became excessively long due to redundant elements within nested structures. However, expert feedback was crucial in addressing these issues, positively shaping model behavior, and guiding it toward producing more accurate and concise schemas. This highlights the efficacy of incorporating domain expert feedback in the workflow of \textsc{schema-miner}. \textbf{4. Schema Property Data Types and Units.} The LLM was effective at assigning accurate data types and units to properties, such as temperature in degrees Celsius, pressure in pascals, dosing time in seconds, thickness in nanometers, and growth per cycle in nanometers/cycle. However, some errors occurred. For instance, in the \href{https://github.com/sciknoworg/schema-miner/blob/main/results/stage-1/simulation-schema/gpt-4o.json}{simulation schema} generated by GPT-4o, the units for reaction rate were incorrectly assigned as atoms/second instead of moles/second. Additionally, the LLM often added unnecessary boolean properties, which confused domain experts about their relevance to the process. These issues could be remedied through continuous expert feedback and introducing more examples of ALD processes. Since \textsc{schema-miner} incorporated expert feedback, this workflow of the LLM and human expert complementing each other is offered as a general recommendation of this work, as it proved to be efficacious in various scenarios in our experiments. \textbf{5. Repetition of Properties.} Domain experts noted repeated properties within the schema, causing confusion. For example, in the \href{https://github.com/sciknoworg/schema-miner/blob/main/results/stage-3/simulation-schema/Experiment-4/meta-llama-3.1-8b-instruct.json}{schema} generated by LLama 3.1 (8B), properties such as uniformity, roughness, density, temperature profile, and chemical composition were duplicated in both \texttt{observables.filmProperties} and \texttt{experimentalResults.results.filmProperties}, with nearly identical structures and descriptions. The issue was corrected by including specific instructions in the prompt for the LLM to avoid repeating properties based on domain expert feedback. The experiments were rerun to better effect.
\textbf{6. Effect of Stage 1.} In both experimental and simulation use cases, the process specification document provided a strong foundation for LLMs to list and structure ALD process properties. The \href{https://github.com/sciknoworg/schema-miner/blob/main/data/stage-1/ALD-Process-Development.pdf}{experimental} specification document included properties like precursor and co-reactant selection, thickness control, saturation, and material properties. Similarly, the \href{https://github.com/sciknoworg/schema-miner/blob/main/data/stage-1/ALD-E_Simulation-Parameters-Observables-List.pdf}{simulation} document covered properties such as growth rate, surface desorption, decomposition, and binding. Using these documents and the pre-trained knowledge of LLMs, initial schemas were generated. The schemas organized relevant information into nested objects, creating semantic clusters of related properties. For example, in the experimental case, precursor and co-reactant were grouped as reactants, while temperature, pressure, and cycle details were grouped as process conditions. This proved to be a strong base of basic properties to build upon for later stages 2 and 3, looking at scientific literature, allowing the schemas to evolve into more advanced forms. \textbf{7. Effect of Larger Research Paper Collection in Stage 3.} In stage 3, a broader collection of research papers, including ALD review papers, was used. This caused some models to deviate from the original schema by incorporating overly specific properties tied to individual papers. GPT-4-turbo was particularly affected in the simulation use case, showing significant divergence from the schema created in stage 2. In contrast, GPT-4o and LLama 3.1 (8B) remained relatively consistent. \textbf{8. Effect of the Different Feedback Methods.} In evaluating feedback methods, descriptive feedback, schema editing, and their combination were tested. In the simulation use case, descriptive feedback guided models to include missing details and correct inaccuracies. For example, experts noted the need for methodological details (e.g., timesteps for MD simulations or functionals for DFT) and distinctions between simulation and experimental findings, prompting improvements to align schemas with expert expectations. Edited schemas allowed experts to directly correct errors, such as removing invalid properties (e.g., substrate velocity) and refining groupings (e.g., limiting reactor conditions to pressure, carrier gas flow, precursor flow, and gap distance). The most effective approach combined both methods, offering clarity on missing details through descriptive feedback and precise corrections through edits. This comprehensive strategy improved schema groupings, added domain-relevant properties, and reduced inconsistencies, demonstrating its efficacy in aligning schemas with expert standards. Note that we also tested the no feedback setting in experiment 4. Experts observed that schemas became overly specific to individual research papers, as no restrictions were imposed on the model’s output, leading to derailment. \textbf{9. Overall LLM Performance in \textsc{schema-miner} across Stages.} In stage 1, GPT-4o and LLama 3.1 (8B) demonstrated the highest structural and semantic coherence in the experimental use case, while LLama 3.1 (8B) excelled in the simulation use case with superior classification and extraction. In stage 2, domain experts observed that more complex LLMs tended to deviate from feedback, but GPT-4-turbo performed best for the experimental use case, identifying key properties without adding irrelevant details. For the simulation use case, GPT-4o produced the most structured and detailed schema. The best results in this stage were achieved when both feedback formats—descriptive and schema edits—were incorporated, providing clear guidance for corrections. In stage 3, using a larger collection of scientific papers, GPT-4o and LLama 3.1 (8B) outperformed GPT-4-turbo in structuring and identifying key properties for both use cases, while GPT-4-turbo's schemas lacked detail, included irrelevant properties, and exhibited poor structuring. The \href{https://orkg.org/template/R796110}{final schema} for the experimental use case (see \autoref{fig:experimental-schema-uml}), produced by GPT-4o, has four levels of nesting with top-level properties as ALD system, reactant selection, process parameters, and material properties, further detailing nested aspects like reactor and thickness control. For the simulation use case, the \href{https://orkg.org/template/R1364068}{final schema}, selected from GPT-4o and LLama 3.1 (8B) outputs, includes three levels of nesting with top-level properties like simulation parameters, materials, growth rate, and reactor conditions.

\section{Discussion}

While our evaluation discussion in this paper focused on the ALD use case in materials science, \textsc{schema-miner} is inherently designed to be domain- and use-case-agnostic, broadly applicable across any scientific discipline. Its modular architecture allows users to provide domain specifications and relevant literature corpora, making it highly adaptable to new use cases. In the life sciences, for example, schema-miner could model workflows such as artificial enzyme engineering, where process parameters include substrate binding affinities, catalytic turnover rates, and amino acid modifications—all commonly embedded in diverse reporting styles across biochemical literature. In DNA autocatalysis research, relevant parameters like reaction rates, temperature profiles, and nucleotide sequences are often reported heterogeneously, necessitating automated schema generation to consolidate process knowledge. Similarly, for metabolic reaction networks, \textsc{schema-miner} can help extract structured information about metabolite concentrations, flux rates, enzyme activity levels, and regulatory feedback loops from pathway modeling studies. As additional domain examples, in environmental science, use cases such as redox chemistry in soils involve parameters like redox potential, microbial community composition, electron donor/acceptor availability, and mineralogy—typically scattered across field study papers, lab analyses, and simulation results. Manually standardizing this information into a schema is prohibitively time-consuming. A tool like \textsc{schema-miner}, operating over a large, heterogeneous corpus, helps to surface common patterns and abstract them into generalized, semantically meaningful schemas, enabling integration across studies. In engineering sciences, as another domain, \textsc{schema-miner} can be applied to fluid dynamics simulations by extracting key parameters such as flow velocity, pressure gradients, turbulence models, boundary conditions, and mesh resolution from computational studies and experimental setups. Even in social sciences, \textsc{schema-miner} can be used to develop structured representations of qualitative research protocols by extracting coding themes, participant demographics, and interview formats from ethnographic or survey-based studies. 

In all these examples, the richness, variability, and scale of scientific reporting make automated schema discovery essential. Schema-miner’s human-in-the-loop design enables domain experts to guide schema evolution while leveraging the scalability of LLMs to parse and structure large, unstructured corpora. This approach bridges the gap between free-text scientific discourse and machine-actionable knowledge, supporting the creation of semantically coherent schemas that standardize information and enable downstream applications such as KG construction and cross-disciplinary integration.









\section{Conclusion and Future Work}

We presented \textsc{schema-miner}, a tool for LLM-assisted schema discovery, outlining its workflow and capabilities. Applied to a complex scientific domain, our experiments demonstrated the effectiveness of combining LLMs with expert feedback in a structured human-in-the-loop process. A \href{https://github.com/sciknoworg/schema-miner/blob/main/MAINTENANCE.md}{maintenance plan} has been released, and we invite \href{https://github.com/sciknoworg/schema-miner/blob/main/CONTRIBUTING.md}{community contributions} to support ongoing development and feature expansion.
Future work could explore ensemble approaches that combine the strengths of multiple LLMs—for example, integrating one model’s structural precision with another’s domain-specific semantics—to generate more robust, generalizable schemas. This could further optimize schema discovery for scientific knowledge representation and AI-ready data modeling.


\begin{credits}
\subsubsection{\ackname} This work was supported by the AWASES project (funded by Merck and Intel), the KISSKI AI Service Center (BMBF, Grant ID: 01IS22093C) and \href{https://scinext-project.github.io/}{SCINEXT} (BMBF, Grant ID: 01IS22070).


\subsubsection{\discintname} The authors have no competing interests to declare that are relevant to the content of this article.
\end{credits}



%
%
%
\bibliographystyle{splncs04}
\bibliography{mybibliography}
\end{document}